\documentclass[11pt,a4paper,onecolumn]{AweAI}

\usepackage[numbers,sort&compress,square]{natbib}
\usepackage{adjustbox}
\usepackage{float}
\usepackage{xurl}
\newcolumntype{L}[1]{>{\raggedright\arraybackslash}p{#1}}
\newcolumntype{Y}{>{\raggedright\arraybackslash}X}
\graphicspath{{fig/}}
\title{A Survey of Large Language Models for Law: Task Capabilities, Authority Grounding, and System Governance}
\hypersetup{
    pdftitle={A Survey of Large Language Models for Law: Task Capabilities, Authority Grounding, and System Governance},
    pdfauthor={Zhongxiang Sun, Xuesheng Li, Weijie Yu, Jun Xu},
    pdfsubject={A critical narrative survey of authority-grounded legal AI systems},
    pdfkeywords={large language models, legal AI, authority grounding, retrieval-augmented generation, evaluation, governance}
}
\pdfcatalog{/Lang (en-US)}
\renewcommand{\today}{}
\keywords{large language models; legal artificial intelligence; legal natural language processing; legal information retrieval; trustworthy artificial intelligence; artificial intelligence governance}
\makeatletter
\renewcommand{\abscontent}{%
    \noindent\parbox{\linewidth}{%
        \absfont\theabstract
        \vskip0.8em
        {\keywordstitlefont Keywords:} {\absfont\@keywords}%
    }%
}
\fancypagestyle{firststyle}{
    \fancyhf{}
    \fancyfoot[L]{\footerfont\textsuperscript{*}Corresponding author: \href{mailto:yu@uibe.edu.cn}{yu@uibe.edu.cn}}
    \fancyfoot[R]{\footerfont\thepage}
    
}
\makeatother

\author[1]{Zhongxiang Sun}
\author[2]{Xuesheng Li}
\author[2]{Weijie Yu\textsuperscript{*}}
\author[1]{Jun Xu}
\affil[1]{Gaoling School of Artificial Intelligence, Renmin University of China, Beijing 100872, China}
\affil[2]{School of Artificial Intelligence and Data Science, University of International Business and Economics, Beijing 100029, China}

\begin{abstract}
Large language models are moving legal artificial intelligence beyond isolated-task evaluation toward composite systems that integrate domain adaptation, retrieval-augmented generation, knowledge organization, agents, and human collaboration. Existing research is often organized around task performance, model mechanisms, or risk governance, with limited integration across these perspectives. This fragmentation makes it difficult to determine when a legal output can justifiably be relied upon. Legal conclusions depend on textual relevance and proposition-level support, as well as source hierarchy and binding force, jurisdiction, temporal validity, and procedural posture. We propose the \emph{authority-grounded legal AI system} as an organizing framework connecting these conditions with system design and evaluation through three non-exclusive dimensions: legal tasks, system mechanisms, and assurance evidence. We synthesize research available through July 2026 and construct a structured evidence map of 327 technical studies from the four complete years 2022--2025. The complete-year window applies to the quantitative snapshot; relevant 2026 studies also inform the narrative synthesis. We compare domain adaptation, retrieval augmentation, knowledge structures, and agents according to how they provide access to authoritative sources, assess legal applicability, verify proposition support, and allocate professional responsibility. The synthesis shows that task accuracy, retrieval relevance, citation faithfulness, and professional usability describe different system properties and cannot be reduced to a single measure of legal reliability. Retrieval of relevant material also does not establish authority grounding. These distinctions motivate research on temporal and jurisdictional robustness, professional workflows, corpus and source governance, abstention and human escalation, and evidence-package reporting. The framework provides a common conceptual structure and evaluation vocabulary for comparing heterogeneous research and jointly examining sources, applicability, supporting evidence, and system responsibility.
\end{abstract}

\begin{document}
\maketitle

\section{Introduction}

Large language models (LLMs) have changed both the capabilities and the system boundary of legal artificial intelligence. Early studies asked whether a general-purpose model could solve an individual task, such as judgment prediction, statutory reasoning, legal entailment, or a bar-exam question \citep{trautmann2022legal,blair2023can,yu2022legal,choi2023chatgpt}. The field now includes retrieval-augmented question answering, law-specific adaptation, multilingual and jurisdiction-specific benchmarks, legal agents, and explicit evaluations of grounding, citation, and abstention \citep{lawbenchbenchmarkingknowledge2023,legalbenchragbenchmark2024,legalagentbenchevaluatingllm2024,measuringfaithfulnessabstention2025}. This expansion creates a conceptual mismatch: a task-centric map can describe what systems attempt, but not the evidence needed to justify reliance on their outputs.

\subsection{Research Background and Core Question}

The main argument of this survey is that LLMs for law should be examined not only as models that perform legal NLP tasks but also as components of \emph{authority-grounded legal AI systems}. We use this proposed term for systems whose answers are assessed against legal sources, jurisdiction and time, an inspectable workflow, and evidence beyond fluency. This is an organizing choice rather than a new task category: it shifts analysis from the model and its answer to the system, its authority chain, and its operating conditions. The framework provides a common lens for prompting, domain adaptation, retrieval, citation grounding, agents, and evaluation without claiming that the literature has already validated all of its requirements.

Law is a distinctive domain for LLM research. Legal-NLP surveys and corpora document long and structured documents, specialized terminology, multiple jurisdictions, and time-sensitive sources \citep{katz2023natural,naturalprocessingdomain2024,multilegalpilegbmultilingual2023}. Studies of legal research tools, case-based argument generation, and case-summary bias provide direct evidence for narrower failures involving citations, faithfulness, abstention, and biased source or summary construction \citep{hallucinationfreeassessingreliability2024,measuringfaithfulnessabstention2025,questioningbiasescase2023}. These results motivate asking legal systems to expose authority, jurisdiction, date, and reasoning support; they do not establish that every legal LLM exhibits every failure.

Reliance on a legal output requires separate answers to three questions: whether the system found material relevant to the issue, what legal status that material has in the target legal system, and whether it supports the specific propositions in the output. Studies often measure only one of these properties, such as task correctness, retrieval relevance, or citation support. The same term, ``reliability,'' consequently refers to different evidence. We distinguish the object of measurement, scope of applicability, and limits of generalization when comparing legal LLM research.

\begin{figure}[t]
\centering
\includegraphics[width=\textwidth]{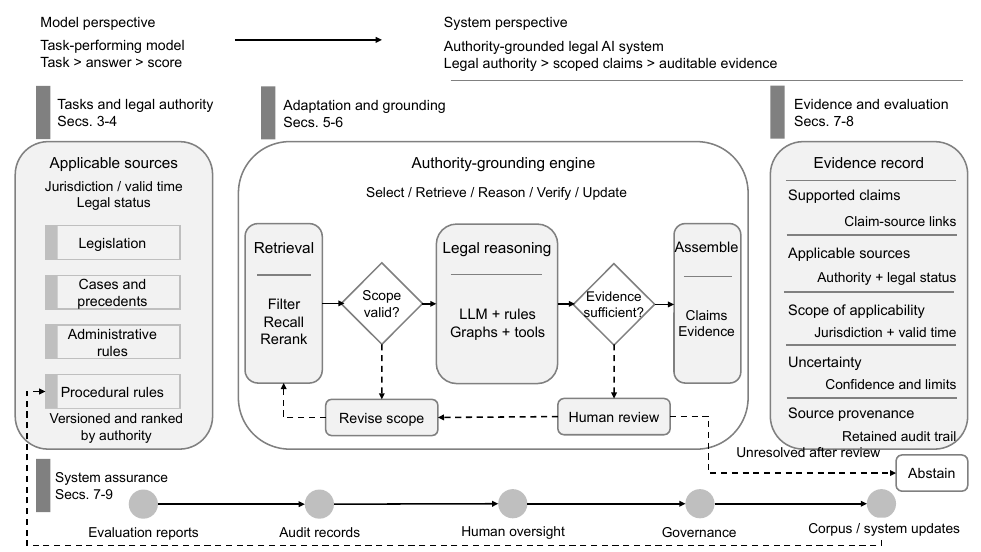}
\caption{The authority-grounded legal AI system framework and reading path through the survey. Solid lines show the main processing and evidence flow; dashed lines show scope revision, human review, and abstention. Labels and structure distinguish the grayscale regions. Re-retrieval, human review, and abstention are proposed design requirements rather than features implemented by all existing systems.}
\label{fig:main-taxonomy}
\end{figure}

\subsection{From Task Maps to System Evidence}

Early LLM-and-law studies often evaluated whether a general-purpose model could perform a legal task under zero-shot, few-shot, or prompt-engineered settings \citep{blair2023can,yu2022legal}. Alongside this task-oriented line, recent work examines components of larger systems: retrieval pipelines, collaborations between general and domain models, multi-agent workflows, and benchmark harnesses for retrieval, faithfulness, or abstention \citep{bladeenhancingblack2024,agentscourtbuildingjudicial2024,legalbenchragbenchmark2024,measuringfaithfulnessabstention2025}. Legal corpora such as MultiLegalPile provide training resources rather than evidence of an integrated LLM system \citep{multilegalpilegbmultilingual2023}. Figure~\ref{fig:main-taxonomy} presents the authors' proposed map of these components, from legal tasks and authority through adaptation and grounding to evaluation and governance.

The literature contains implementations based on prompt-only inference, domain tuning, retrieval, tool use, multi-agent orchestration, and hybrid symbolic components. We use these alternatives to motivate three linked review questions: what legal capability is claimed, what mechanism produces it, and what assurance evidence is reported for the stated jurisdiction, time, and workflow? This is the survey's analytical choice rather than a claim that task taxonomies have ceased to be useful.

This survey takes a system-oriented and evaluation-oriented view. We do not treat ``law'' as merely another domain label attached to generic NLP. Instead, we view legal LLM research as the interaction of three constraints: (i) legal tasks require authority-sensitive reasoning; (ii) LLM systems require mechanisms for adaptation, retrieval, and control; and (iii) deployment requires evidence of grounding, robustness, and professional usability. This perspective lets us compare papers that would otherwise appear in separate clusters, including legal case retrieval, legal-domain pretraining, legal hallucination detection, and legal-agent benchmarking.

General categories of applications, resources, and risks are inherited from prior reviews. Our additions are a structured public-source search, the task--mechanism--assurance taxonomy, a definition of authority grounding, and an integrated analysis of retrieval-augmented generation (RAG), knowledge structures, agents, evaluation, and governance. The evidence map covers 381 screened records and 327 included technical studies. Its title-and-abstract codes support reproducibility and literature navigation; exploratory quantitative panels appear in the supplementary materials. The synthesis asks what evidence each study supplies, where that evidence ends, and whether different evidence can be combined within a common system boundary.

Authority grounding is not a synonym for legal RAG, citation generation, accountable AI, or legal agents. RAG supplies candidate sources but does not by itself establish that they are binding or persuasive, current, applicable to the relevant jurisdiction, or supportive of a particular proposition; citation grounding can test source--claim support without resolving those legal-status questions; agents orchestrate tools without automatically assigning responsibility. We therefore propose an operational integration contract with four parts: source-applicability checks; proposition-level claim--source links; explicit re-retrieval, escalation, or abstention rules; and an assurance record for evaluation, review, and updating. These parts are design and reporting requirements proposed by this survey, not an empirically certified deployment standard.

Our main contributions are:
\begin{itemize}[leftmargin=1.5em]
    \item A definition of authority grounding and a taxonomy connecting task capability, system mechanisms, and assurance evidence, with explicit distinctions among relevance, proposition support, and legal applicability.
    \item A comparison of how prompting, domain adaptation, retrieval, knowledge structures, and agents change the authority chain, updating path, verification burden, and failure surface, supported by an inspectable evidence map of 327 technical studies.
    \item An evaluation and governance agenda connecting source faithfulness, jurisdiction, valid time, abstention, human responsibility, and operational monitoring to testable system boundaries.
\end{itemize}

\subsection{Organization of the Survey}
Section~2 describes the review scope, search procedure, and taxonomy. Section~3 introduces legal tasks, texts, reasoning, and authority. Sections~4--6 examine applications, domain adaptation, and retrieval and agents. Section~7 compares benchmarks and system evaluation, and Section~8 discusses governance across models, sources, professionals, institutions, and deployment. Section~9 develops the research agenda, Section~10 sets out the limitations, and Section~11 concludes.

\section{Scope and Taxonomy}

\subsection{Scope, Evidence Boundaries, and Review Method}

We focus on work where LLMs are used for legal-domain tasks, legal reasoning, legal-domain adaptation, legal evaluation, or legal-risk analysis. Generic NLP work is included only when it directly illuminates a mechanism used in legal AI, while traditional legal NLP work is used as background when it defines an influential dataset, task formulation, or institutional constraint.

This article is a critical narrative survey rather than a systematic review, meta-analysis, or bibliometric study. Its purpose is to connect research threads that are often discussed separately and to propose a system-level account of legal reliability. The conceptual synthesis is supplemented by a structured public-source search and paper-level evidence map, but the sample is not treated as an exhaustive census, a prevalence estimate, or an empirical validation of the proposed framework. Individual technical claims remain tied to the cited studies that support them.

The synthesis covers relevant work available through 17 July 2026. To make the quantitative snapshot inspectable and comparable across complete years, we searched arXiv and OpenAlex for records published in 2022--2025. The arXiv query required \texttt{legal}, \texttt{law}, \texttt{judicial}, \texttt{statute}, \texttt{court}, or \texttt{jurisdiction} in the title and \texttt{``large language model''}, \texttt{LLM}, or \texttt{ChatGPT} in the indexed record. The OpenAlex query used full-text search for \texttt{``large language model'' law legal}; only the first 200 relevance-ranked records were retrieved. The completed arXiv and OpenAlex exports contained 969 raw API rows. Deduplication by arXiv identifier, DOI, and normalized title produced 915 unique records and 929 retained source memberships.

Record-level rules removed retractions, paratext, non-legal uses of words such as ``law'' or ``court,'' and records without an explicit LLM and title-level legal-domain signal; partial-year 2026 records were held outside the quantitative window. The narrative synthesis applies the same relevance and evidence standards to relevant work from every year; the frozen window limits only the reported quantitative counts. No publisher-wide exclusion was applied. A sensitivity check identified 16 MDPI-hosted records, all of which remained outside the screening pool under the same content rules used for other publishers. These filters produced 381 candidates for paper-level screening.

All 381 candidates have explicit title-and-abstract decisions, reason codes, evidence bases, and screening-pass identifiers. The single-reviewer process retained 327 primary technical records, treated 33 surveys, governance papers, and mechanism-adjacent works as background, excluded 19 out-of-scope records, and kept two unresolved records outside every quantitative artifact. Retrieval provenance and publication status are recorded separately: 313 included records were returned only by the arXiv search, 11 by both arXiv and OpenAlex, and three only by OpenAlex. These source memberships do not indicate whether a work has a formal venue version. Record-level verification identified 191 formally published versions, one accepted or in-press version, 134 records for which only a preprint or working-paper version was confirmed, and one unresolved publication-status record; version updates do not change sample membership. Task and mechanism labels are non-exclusive automated title-and-abstract signals. All 114 triggered assurance pairs received a conservative single-reviewer re-adjudication, but no independent second coder or inter-coder agreement is claimed. The corrected plots are therefore exploratory Supplementary Data rather than main-text evidence.

A study enters the core discussion when it contributes to at least one part of the system view developed here: it evaluates an LLM on a legal task; introduces a legal-domain model, dataset, benchmark, retrieval system, or agent; studies a legally meaningful failure such as hallucinated authority, jurisdictional mismatch, temporal invalidity, privacy leakage, or biased legal output; or proposes a mechanism for legal grounding, reasoning, adaptation, verification, or governance. Generic work on LLM safety, RAG, or agents is discussed only when it clarifies a mechanism that is instantiated in legal AI. These principles define the intellectual scope of the survey, not a claim of database-complete selection.

\subsection{Related Surveys and Analytical Questions}

Prior surveys provide foundations for legal NLP, deep learning in law, judgment prediction, and LLM applications \citep{chalkidis2019deep,cui2022survey,dias2022state,katz2023natural,naturalprocessingdomain2024}. Reviews published in 2025 offer complementary application-centered, model-and-resource, and dual-perspective taxonomies \citep{siino2025exploring,hou2025legalaissurvey,shao2025duallenssurvey}. A 2026 review further focuses on legal-agent architectures, workflows, applications, and risks \citep{agentslawtaxonomyapplications2026}. Building on these task, resource, and risk categories, we add a structured public-source search, a task--mechanism--assurance taxonomy, and an authority-grounded system perspective connecting RAG, agents, evaluation, and governance. Table~\ref{tab:survey-comparison} describes how the reviews relate to one another without scoring their completeness.

\begin{table}[!htbp]
\centering
\footnotesize
\setlength{\tabcolsep}{3.5pt}
\caption{Positioning relative to prior surveys. The comparison reports each work's stated evidence base and organizing scope; it is not a completeness score or coverage ranking.}
\label{tab:survey-comparison}
\begin{tabular}{L{0.20\linewidth} L{0.22\linewidth} L{0.27\linewidth} L{0.23\linewidth}}
\toprule
Survey/year & Evidence base & Organizing scope & Relation to this review \\
\midrule
\citet{chalkidis2019deep} & legal deep-learning literature & representations, architectures, and legal applications & methodological foundation before the LLM-centered literature \\
\citet{cui2022survey} & judgment-prediction studies & datasets, methods, and evaluation for legal judgment prediction & task-specific foundation for capability analysis \\
\citet{dias2022state} & broad AI-and-law literature & methods and applications across legal domains & broad legal-AI context \\
\citet{katz2023natural} & legal NLP research and resources & tasks, datasets, models, and institutional setting & comprehensive legal-NLP foundation \\
\citet{naturalprocessingdomain2024} & legal NLP literature & tasks, datasets, resources, and methods & updated domain-level foundation \\
\citet{surveycriticalsocietaldomains2024} & cross-domain LLM literature & risks and applications in finance, health, and law & comparative cross-domain risk context \\
\citet{siino2025exploring} & database search and snowballing; 61 studies from 2017--2023 & legal document analysis, prediction, contract review, and related applications & structured application-focused literature review \\
\citet{shao2025duallenssurvey} & legal-LLM literature and public paper index & legal roles, NLP subtasks, argumentation, technical advances, and governance & contemporary dual-lens taxonomy and governance synthesis \\
\citet{hou2025legalaissurvey} & curated catalog of models, frameworks, benchmarks, and datasets & legal LLMs, task frameworks, resources, and challenges & contemporary model-and-resource map \\
\citet{agentslawtaxonomyapplications2026} & legal-agent literature & agent architectures, workflows, applications, and risks & agent-specific system perspective \\
\midrule
This review & critical narrative synthesis plus a 327-record public-source evidence map & task capability, system mechanism, legal authority, assurance, and governance & authority-grounded system as the proposed unit of analysis \\
\bottomrule
\end{tabular}
\end{table}

The synthesis follows six questions across technical threads: what legal capability is claimed; what model or system mechanism produces it; which legal sources ground the output; how jurisdiction and time constrain applicability; what evidence supports reliability; and what risks, human controls, or governance obligations remain. These questions provide a common reading frame for heterogeneous studies. They are not a quantitative coding protocol and do not turn the cited literature into a statistical sample.

\subsection{The Three Axes and Units of Evidence}

We organize the area along three axes:
\begin{enumerate}[leftmargin=1.5em]
    \item \textbf{Legal task layer}: judgment prediction, statutory reasoning, case retrieval, legal QA, summarization, drafting, information extraction, compliance, education, dispute resolution, and access-to-justice assistance.
    \item \textbf{System mechanism layer}: prompting, chain-of-thought and structured reasoning, domain pretraining, supervised/instruction tuning, retrieval-augmented generation, knowledge graphs, tool use, multi-agent workflows, and neuro-symbolic constraints.
    \item \textbf{Assurance and evaluation layer}: factual grounding, citation faithfulness, jurisdictional robustness, temporal validity, privacy, fairness, explainability, abstention, and professional usability.
\end{enumerate}

\begin{figure}[t]
\centering
\includegraphics[width=\textwidth]{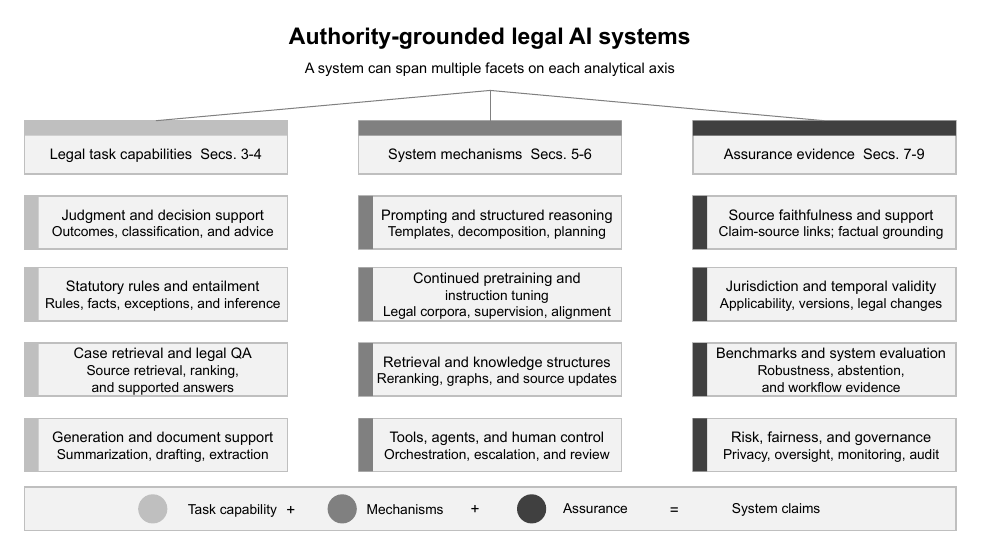}
\caption{The task--mechanism--assurance taxonomy of legal LLM research. The axes are non-exclusive facets: a system can involve several tasks, mechanisms, and assurance dimensions. Connections indicate possible membership, not frequency or causation.}
\label{fig:faceted-taxonomy}
\end{figure}

Figure~\ref{fig:faceted-taxonomy} renders these layers as non-exclusive facets, not as a mutually exclusive hierarchy. A single study or deployed system can therefore connect several task, mechanism, and assurance categories.

Table~\ref{tab:representative-studies} compares representative resources, benchmarks, prototypes, and deployment-adjacent evaluations at a common system boundary. The comparison deliberately separates what a study releases or tests from what it does not establish. In particular, a corpus is not an integrated legal LLM, a retrieval benchmark is not an end-to-end legal service, a temporal event-ordering benchmark is not a law-versioning test, and a professional-tool audit is not a controlled workflow trial.

\begin{table}[!htbp]
\centering
\footnotesize
\setlength{\tabcolsep}{3.5pt}
\caption{Representative studies viewed at the system level. ``Evidence status'' describes what the cited study evaluates or releases; it does not imply professional deployment or legal validity beyond the stated setting.}
\label{tab:representative-studies}
\begin{tabular}{L{0.16\linewidth} L{0.20\linewidth} L{0.20\linewidth} L{0.21\linewidth} L{0.15\linewidth}}
\toprule
Study/system & Task and jurisdiction & Mechanism & Authority or assurance evidence & Evidence status \\
\midrule
LawBench \citep{lawbenchbenchmarkingknowledge2023} & legal knowledge and reasoning; Chinese law & benchmark across knowledge, understanding, and application tasks & task performance across model families & released capability benchmark; not a deployment study \\
MultiLegalPile \citep{multilegalpilegbmultilingual2023} & legal corpus; 24 languages and 17 jurisdictions & corpus construction and domain pretraining & provenance, language, jurisdiction, and license metadata & released resource and models; not a generative legal system \\
AgentsCourt \citep{agentscourtbuildingjudicial2024} & simulated judicial decision making; Chinese cases & court-debate agents, legal retrieval, and decision refinement & benchmarked final decisions and article generation & research prototype evaluated in simulation \\
LegalBench-RAG \citep{legalbenchragbenchmark2024} & legal passage retrieval; heterogeneous legal sources & retrieval benchmark for RAG pipelines & expert-annotated minimal relevant passages & released retrieval benchmark; generation and deployment excluded \\
LegalAgentBench \citep{legalagentbenchevaluatingllm2024} & multi-step legal tasks; Chinese legal scenarios & agents interacting with 37 external tools & final success plus intermediate progress signals & released agent benchmark; no professional workflow trial \\
Legal research-tool audit \citep{hallucinationfreeassessingreliability2024} & commercial legal research tools; U.S. legal research & proprietary RAG-based products & preregistered hallucination, responsiveness, and accuracy evaluation & deployment-adjacent tool audit; not a controlled user study \\
Questioning Biases \citep{questioningbiasescase2023} & case-judgment summarization & dataset and model comparison & analysis of bias in legal summaries and source datasets & empirical research study; no deployment claim \\
LexTime \citep{lextimebenchmarktemporalordering2025} & event ordering in U.S. federal complaints & temporal-relation benchmark & annotated event pairs and ordering accuracy & released task benchmark; does not evaluate law versioning \\
Faithfulness and abstention \citep{measuringfaithfulnessabstention2025} & three-ply case-based argument generation & automated factor extraction and comparison & faithfulness, factor use, and instructed abstention & research evaluation pipeline; no deployment claim \\
TRIDENT \citep{hui2025tridentbenchmarkingllmsafety} & professional safety in law, finance, and medicine & principle-based safety benchmark & domain-specific ethical and compliance scenarios & cross-domain benchmark; legal deployment not evaluated \\
\bottomrule
\end{tabular}
\end{table}

A single-axis taxonomy hides important differences. If we classify only by task, legal QA systems that use prompt-only inference, retrieval grounding, fine-tuned legal LLMs, and agentic tool use would be grouped together despite having different assumptions and risks. If we classify only by mechanism, domain adaptation papers and RAG papers may appear unrelated even when both address the same underlying need: connecting model outputs to legal authority. If we classify only by benchmark, we miss whether an evaluation measures legal knowledge, legal reasoning, source faithfulness, jurisdictional transfer, or professional usefulness. The three-axis view keeps these distinctions visible throughout the survey.

\section{Foundations: Legal Tasks, Legal Text, and Legal Reasoning}

LLM-based legal AI inherits task definitions from legal NLP, but the introduction of generative and instruction-following models changes how these tasks should be analyzed. A classifier for legal judgment prediction can be evaluated with labels; a legal assistant that drafts an answer must be evaluated for source grounding, jurisdictional scope, factual assumptions, and downstream human use. This section establishes the foundations needed for the taxonomy used in the rest of the survey.

\subsection{Tasks and Legal Texts}

\begin{table}[!htbp]
\centering
\footnotesize
\caption{Task families and their assurance obligations. As outputs become more open-ended, evaluation shifts from label correctness toward source support, scope, and workflow control.}
\label{tab:legal-task-taxonomy}
\begin{adjustbox}{max width=\linewidth}
\begin{tabular}{L{0.18\linewidth} L{0.24\linewidth} L{0.24\linewidth} L{0.24\linewidth}}
\toprule
Task family & Typical input & Expected output & Core evaluation question \\
\midrule
Judgment prediction & Facts, charges, claims, prior cases & charge, article, term, disposition & Does the model identify legally relevant facts and applicable rules? \\
Statutory reasoning & Statute, regulation, question, fact pattern & answer and rule application & Does the answer respect statutory conditions and exceptions? \\
Case retrieval & query case or legal issue & ranked cases or passages & Are retrieved authorities relevant, controlling, and explainable? \\
Legal QA & user question plus legal sources & grounded answer with citations & Is every material statement supported by valid authority? \\
Summarization & long judgment, contract, policy, or brief & concise summary or issue list & Are holdings, facts, and procedural posture preserved? \\
Drafting and review & instructions, facts, templates, documents & clause, memo, complaint, contract review & Is generated text legally coherent and safe for review? \\
Compliance and extraction & regulation, policy, form, evidence & structured obligations or entities & Are obligations, conditions, and exceptions captured? \\
Agents and workflows & multi-step task or dialogue & tool-mediated process and final answer & Are intermediate steps auditable and controllable? \\
\bottomrule
\end{tabular}
\end{adjustbox}
\end{table}

Legal AI tasks range from classification and retrieval to open-ended reasoning and generation. Traditional benchmark tasks include legal judgment prediction, legal entailment, case retrieval, legal question answering, and legal summarization \citep{xiao2018cail2018,zheng2021does,ma2021lecard,katz2023natural,naturalprocessingdomain2024}. LLM-era tasks increasingly include legal consultation, document drafting, compliance checking, legal argument generation, and interactive legal agents \citep{lawbenchbenchmarkingknowledge2023,legalagentbenchevaluatingllm2024,legalevalqnew2025}. Table~\ref{tab:legal-task-taxonomy} summarizes task families and the evaluation questions they raise.

Legal corpora differ from many web-text corpora because they contain long documents, specialized terminology, citations, hierarchical provisions, procedural context, and jurisdiction-specific meanings \citep{katz2023natural,naturalprocessingdomain2024,multilegalpilegbmultilingual2023}. A statute may define a rule through nested conditions and exceptions; a judicial opinion may combine facts, procedural history, holdings, dicta, and citations; a contract may encode obligations, rights, conditions, and remedies across distant clauses. These properties make long-context modeling and retrieval design central to legal LLM systems. Passage-level retrieval can lose relationships across distant provisions, so segmentation and retrieval choices must be evaluated for the task rather than assumed to preserve legal context \citep{enhancingdocumentretrievalmulti2024,legalbenchragbenchmark2024}.

\subsection{Defining Authority Grounding}

Legal authority is not captured by document relevance alone. Legal systems distinguish material sources of law from commentary, and the status of a source depends on the practices and institutions of the relevant legal community \citep{bell2018sources,maccormick1994legalreasoning}. For system design, this requires at least six explicit attributes. First, the source type should distinguish constitutions, legislation, regulations, judicial decisions, administrative materials, contracts, and secondary commentary. Second, the system should record whether an authority is binding, persuasive, or otherwise informative for the relevant decision maker. Third, hierarchy and jurisdiction determine which institution's materials can govern the question. Fourth, valid time and subsequent treatment matter because provisions may be amended and decisions may be overruled, limited, or distinguished. Fifth, procedural posture and the distinction between holding and dicta affect how a judicial passage can be used. Sixth, conflicting sources or interpretations should remain visible rather than being silently collapsed into a single answer. These attributes are a cross-system checklist proposed for this survey; their legal effect must be specified for the jurisdiction under study.

We distinguish citing legal text from establishing a basis for legal reliance through the following definition. A generated proposition is \emph{authority grounded} when the system can record and test the applicability of a candidate source under specified operating conditions, and establish an inspectable support relation between the proposition and that source. A relevant source fails this definition if it is no longer valid, belongs to an inapplicable jurisdiction, or does not substantively support the proposition. This analytical definition separates legal applicability from textual support as independently testable conditions; it does not assume that automated methods can fully and independently determine legal effect.

Verification can be divided into three stages. \emph{Retrieval eligibility} checks whether candidate materials come from an admissible collection with identifiable versions. \emph{Legal applicability} checks source hierarchy, binding force, jurisdiction, valid time, and procedural context. \emph{Proposition support} then tests whether the cited passages support the material claims. Retrievers, rules, knowledge graphs, LLM verifiers, and human reviewers may contribute to these checks, but their errors should be reported separately. A source may support a proposition literally while lacking the status or scope required to serve as controlling authority.

Legal reasoning includes statutory interpretation, analogical reasoning over precedents, deontic reasoning over obligations and permissions, temporal reasoning over legal changes, and procedural reasoning over institutional steps. Statutory reasoning often requires checking whether a fact pattern satisfies each element of a rule and whether an exception applies \citep{blair2023can}. Case-based reasoning requires comparing the legally relevant facts and holdings of prior cases rather than merely matching surface text \citep{ma2021lecard,logicrulesexplanations2024}. Deontic reasoning requires distinguishing duties, permissions, prohibitions, exceptions, and sanctions. These reasoning forms are often intertwined in realistic legal tasks.

For many legal tasks, the output is useful only if it can be connected to authority. A legal answer should expose the relevant source, the rule attributed to that source, the facts assumed, and the reasoning step connecting facts to the rule. Citation-grounding and faithfulness studies show why plausible language and a well-formed citation are insufficient when the cited material does not support the generated proposition \citep{measuringfaithfulnessabstention2025,citationgroundingdetecting2026}. Proposition-level support is therefore necessary for the proposed framework, but it is not sufficient: a proposition can be textually supported by a source that is outdated, non-binding, outside the jurisdiction, procedurally inapposite, or contradicted by higher authority.

\subsection{Reasoning Support and System Boundaries}

The proposed unit of analysis is usually a legal AI system rather than a standalone model. Depending on the study, that boundary may include document ingestion, retrieval, reranking, prompting, tool use, generation, citation checking, human review, or logging. We use this component list as an analytical checklist; the relevant boundary and failure locations must be established for each evaluated system.

The foundation for this survey is therefore not only a list of tasks but a shift in the evaluation object. Once legal text is treated as potentially authoritative material whose status must be determined, each task raises questions about how a system connects facts, rules, sources, and institutional context. This motivates the three-axis taxonomy: task capability, system mechanism, and assurance evidence should be analyzed together.

\section{Applications of LLMs in Legal Tasks}

The application literature can be organized by the legal capability being tested. Studies use general-purpose prompting, retrieval, domain models, structured reasoning, and task-specific evaluation. This section uses the distinction between fluent output and evidence-supported output as an analytical lens; it does not assume a measured field-wide gap between the two.

\subsection{Judgment Prediction and Rule-Based Reasoning}

Legal judgment prediction (LJP) asks a system to predict charges, applicable law articles, prison terms, case outcomes, or related dispositions from case facts. Pre-LLM datasets such as CAIL2018 established large-scale formulations of this task \citep{xiao2018cail2018}. LLM-era work explores whether prompting, legal syllogism prompting, discriminative reasoning, precedent enhancement, and domain-model collaboration can improve reasoning over case facts \citep{trautmann2022legal,syllogismpromptingteachingjudgment2023,precedentenhancedjudgment2023,enablingdiscriminativereasoningjudgment2024}. Chinese journal research also extends prediction from single-defendant to multi-defendant cases, using an explicit chain of judgment to represent criminal relationships, sentencing circumstances, applicable articles, charges, and prison terms \citep{lv2025chainjudgment}. LJP is not simply text classification: a robust system should identify legally relevant facts, map them to elements of legal rules, handle missing facts, and avoid learning dataset shortcuts.

Statutory reasoning is a natural testbed for LLMs because the rule is often explicitly provided, but correct application still requires compositional reasoning over conditions, exceptions, and definitions. Studies of GPT-style statutory reasoning, legal prompting, legal textual entailment, and step-by-step legal reasoning show both the promise of prompting and the difficulty of stable legal inference \citep{blair2023can,yu2022legal,nguyen2023blackboxanalysisgptstime,nguyen2024employinglabelmodelschatgpt,mishra2025investigatingshortcomingsllmsstepbystep}. Legal entailment tasks similarly test whether a system can determine whether a legal hypothesis follows from a premise or rule. For high-stakes use, the important question is not only whether the final answer is correct, but whether the reasoning path identifies the controlling rule and the decisive factual elements.

Judgment prediction and statutory reasoning both connect facts, rules, and conclusions, but their scores have different meanings. Prediction often measures accuracy or F-scores over closed labels and a fixed data distribution. Statutory reasoning may supply the rules directly and test how conditions and exceptions combine. Labels extracted automatically from historical judgments can introduce effects of document style, label co-occurrence, or dataset composition, so high scores alone do not establish transferable legal reasoning. Stepwise explanations facilitate inspection but may introduce unsupported facts or rules. The comparable properties are identification of legal elements, factual sufficiency, rule provenance, exception handling, and the ability to locate errors.

\subsection{Case Retrieval and Legal Question Answering}

Legal retrieval asks a system to find relevant cases, statutes, regulations, or passages for a legal issue. LeCaRD introduced a Chinese legal case-retrieval dataset and benchmark setting \citep{ma2021lecard}. LeCaRDv2 expands the setting to 800 queries and 55\,192 candidate cases, incorporating offense characterization, sentencing, and procedure into relevance judgments \citep{li2024lecardv2}. Subsequent LLM-based studies examine query rewriting, content selection, multi-phase retrieval, reranking, explanation generation, and precedent-enhanced prediction \citep{zhou2023boostinglegalcaseretrieval,precedentenhancedjudgment2023,logicrulesexplanations2024,enhancingdocumentretrievalmulti2024,premasiri2025llmbasedembedderspriorcase}. The retrieval setting exposes a central tension: persuasive explanations do not establish that the retrieved material is legally authoritative or factually analogous.

Legal QA systems answer questions from lay users, professionals, or institutional workflows. Work on augmented explanations, privacy-law grounding, legal consultation, proactive dialogue, RAG benchmarks, retrieval--generation interpretation, and faithfulness evaluation studies different ways to add evidence or interaction to free-form answers \citep{explainingconceptsaugmented2023,goldcoingroundingprivacy2024,wu2024knowledgeinfusedlegalwisdomnavigating,chouhan2025answersguidanceproactivedialogue,legalbenchragbenchmark2024,automatinginterpretationretrievalgeneration2025,measuringfaithfulnessabstention2025}. For the proposed framework, source authority, jurisdiction, date, uncertainty, abstention, and clarification should be evaluated when they are material to the intended QA use.

Retrieval and QA require comparisons at individual pipeline stages because a final answer cannot identify where an error arose. Case retrieval first needs a definition of legal relevance: the offense, sentencing, and procedural dimensions in LeCaRDv2 show that similarity is more than semantic distance. QA also involves source selection, rule synthesis, application to facts, and citation presentation. Low recall leaves evidence missing, reranking changes which authorities enter the context, and a generator can form incorrect propositions even when given the correct materials. Candidate coverage, ranking quality, source applicability, proposition support, and abstention therefore need separate reporting.

\subsection{Summarization, Drafting, and Professional Support}

Legal summarization compresses opinions, judgments, statutes, legislative proposals, or contracts into shorter artifacts \citep{katz2023natural,naturalprocessingdomain2024,pont2023legalsummarisationllmsprodigit,howreadyarepre2023,gesnouin2024llamandementlargelanguagemodels,applicabilitygenerativecasejudgement2024,relexedretrievalenhancedsummarization2025,heddaya2024casesummlargescaledatasetlongcontext}. The required content depends on the document and user: case summaries may need facts, holdings, and procedural history, while contract or legislative summaries may emphasize obligations, exceptions, or proposed changes. Studies reporting inconsistencies and hallucinations in case-judgment summaries motivate factual and issue-coverage checks beyond surface overlap \citep{howreadyarepre2023,applicabilitygenerativecasejudgement2024}. A single universal preservation checklist is therefore inappropriate without a task-specific rubric.

Drafting studies examine contracts, legal memos, complaints, clauses, expert-system rules, and regulatory documents \citep{documentsdraftingfinetuned2024,judgebenchmarkingjudgmentdocument2025,legalevalqnew2025}. Educational and professional studies also investigate LLM support for first-draft or assessment tasks \citep{choi2023chatgpt,hargreaves2023words}. These findings do not by themselves establish safe contractual use: generated text can omit clauses, alter obligations, or be misused to create deceptive legal artifacts \citep{mik2026contractualdeepfakeslargelanguage}. Document analysis applications include entity extraction, policy analysis, compliance checking, legal text visualization, regulatory rule extraction, and conversion from text to structured representations \citep{janatian2023textstructureusinglarge,judicialentityextractioncomparative2024,onami2025legalvizlegaltextvisualization,braun2025hiddenstructureimproving,wang2025llmbasedhsecomplianceassessment}. Structured outputs can make validation easier, but the required human review remains task- and risk-dependent.

LLMs have been studied in legal education, law-school assessment, professor support, dispute resolution, moot-court simulation, and professional assistance scenarios \citep{choi2023chatgpt,hargreaves2023words,pettinato2023chatgpt,westermann2023llmediatorgpt4assistedonline,chlapanis2024archimedesauebsemeval2024task5,westermann2024analyzingimageslegaldocuments}. Because these studies test different users and tasks, their findings should not be generalized to autonomous legal advice. They instead motivate workflow-specific evaluation of the assistance provided, the review required, and the person responsible for the resulting decision.

\subsection{Cross-Task Comparison and Evidence Boundaries}

Across applications, we distinguish plausible legal language from output accompanied by evidence about authority, scope, and review. The importance of each property depends on the task and intended use. This analytical distinction motivates the later sections on domain adaptation, retrieval grounding, evaluation, and responsible deployment.

Cross-task comparison requires four dimensions. \emph{Input openness} distinguishes fixed fact descriptions from open consultation. \emph{Output constraints} distinguish closed labels, ranked lists, summaries, and free text. \emph{Dependence on authority} captures the different source-status and temporal requirements of educational explanations and formal legal opinions. \emph{Human responsibility} distinguishes research benchmarks, internal assistance, and systems serving the public directly. Numerical results can be compared directly only when these dimensions and the experimental conditions are compatible. Table~\ref{tab:study-local-evidence} consequently summarizes comparisons within individual studies rather than ranking results across datasets, metrics, or jurisdictions.

\begin{table}[!htbp]
\centering
\footnotesize
\setlength{\tabcolsep}{3pt}
\renewcommand{\arraystretch}{1.12}
\caption{Evidence comparisons within representative studies using shared data, metrics, and evaluation protocols.}
\label{tab:study-local-evidence}
\begin{tabularx}{\linewidth}{@{}L{0.13\linewidth} Y Y Y Y@{}}
\toprule
Study & Comparison setting & Methods or models & Reported evaluation evidence & Limits of inference \\
\midrule
Lawformer \citep{xiao2021lawformer} & CAIL-Long and case retrieval, reading comprehension, and QA data & BERT, RoBERTa, legal RoBERTa, and Lawformer & Classification, regression, retrieval, and QA metrics under the same downstream protocol & Does not fully separate corpus and long-context effects or establish temporal validity \\
LeCaRDv2 \citep{li2024lecardv2} & The same 800 queries and 55\,192 candidate cases & Sparse, dense, and pretrained retrieval baselines in the original study & Retrieval compared under shared relevance labels and ranking protocols & Does not establish open-corpus retrieval, answer quality, or binding force \\
LawBench \citep{lawbenchbenchmarkingknowledge2023} & The same 20 Chinese legal tasks and prompting protocol & 21 general, Chinese-language, and legal models & Results across knowledge memorization, understanding, and application & Scores do not transfer between question sets or establish professional usability \\
LexEval \citep{lexevalcomprehensivechinese2024} & 23 tasks and 14\,150 questions & 38 open and commercial models & Shared capability taxonomy and task metrics & Does not exclude training overlap or cover all regions and legal time points \\
LegalAgent\newline Bench \citep{legalagentbenchevaluatingllm2024} & 17 corpora, 37 tools, and 300 tasks & Eight LLM agents & Final success and intermediate progress & Does not establish efficiency, safety, or accountability in professional workflows \\
Chain of Judgment \citep{lv2025chainjudgment} & The same multi-defendant prediction data and baseline protocol & Judgment-chain generation, comparison strategies, and pretrained baselines & Article, charge, and sentence prediction with analysis of defendant relationships & Does not establish generalization to other case types or generated legal advice \\
\bottomrule
\end{tabularx}
\end{table}

Under the proposed framework, judgment prediction raises rule-element questions, retrieval raises authority-sensitive relevance questions, QA raises jurisdiction and time-scope questions, summarization raises omission-control questions, and drafting raises text-validation questions. Treating these as related assurance obligations is the survey's system-level synthesis, to be tested rather than assumed across applications.

\section{Legal Domain Adaptation and Law LLMs}

LLM adaptation for law includes prompt engineering, retrieval augmentation, continued pretraining on legal corpora, supervised fine-tuning, instruction tuning, synthetic-data generation, and parameter-efficient adaptation. Legal-domain models and resources include early legal pretrained models and recent open-source legal LLMs \citep{zheng2021does,lexgpt2023,chatlaw2023,internlmlaw2024,saullm2024}. Table~\ref{tab:mechanism-map} summarizes major mechanisms and their risks.

\subsection{Prompting and Continued Pretraining}

\begin{table}[!htbp]
\centering
\footnotesize
\caption{Mechanisms, benefits, and failure surfaces. No mechanism alone establishes authority grounding; each changes both the capability gained and the evidence required.}
\label{tab:mechanism-map}
\begin{adjustbox}{max width=\linewidth}
\begin{tabular}{L{0.18\linewidth} L{0.25\linewidth} L{0.25\linewidth} L{0.22\linewidth}}
\toprule
Mechanism & Main benefit & Typical legal use & Main risk \\
\midrule
Prompting & Low-cost adaptation without training & statutory reasoning, exams, explanations & unstable reasoning and prompt sensitivity \\
Domain pretraining & Better legal vocabulary and document familiarity & legal LLMs, embeddings, retrieval encoders & stale knowledge and corpus licensing issues \\
Instruction tuning & Aligns outputs with legal task formats & QA, drafting, summarization & overfitting to synthetic or narrow instructions \\
RAG & Connects outputs to candidate external evidence & legal QA, compliance, citation support & retrieval misses, wrong authority, citation misuse \\
Knowledge graphs & Explicit relation and citation structure & citation grounding, entity/relation extraction & incomplete graph construction and maintenance cost \\
Agents and tools & Decompose long workflows & consultation, simulation, contract review & compounding errors and unclear accountability \\
Neuro-symbolic constraints & Add explicit rules or verification & statutory logic, deontic reasoning, compliance & brittle formalization and coverage gaps \\
\bottomrule
\end{tabular}
\end{adjustbox}
\end{table}

Prompting is the most lightweight form of domain adaptation. Legal prompt engineering and legal prompting work show that task instructions, examples, and reasoning cues can change model behavior \citep{trautmann2022legal,yu2022legal}. Prompting requires no weight update, but the cited studies do not establish source retrieval, authority verification, or stability across operating conditions. For the proposed framework, those properties should therefore be evaluated separately rather than inferred from prompted task performance.

Continued pretraining aims to expose models to legal language, citation patterns, document structure, and domain terminology. Resources such as legal holdings datasets, legal corpora, summarization corpora, citation graphs, and multilingual legal collections support this direction \citep{zheng2021does,lexgpt2023,multilegalpilegbmultilingual2023,naturalprocessingdomain2024,heddaya2024casesummlargescaledatasetlongcontext,lasandi2026sinhalegalbenchmarkcorpusinformation,ovcharov2026automaticconstructionlegalcitation}. Lawformer compares general Chinese BERT and RoBERTa, L-RoBERTa pretrained on the same legal corpus, and the long-document Lawformer under a shared downstream protocol. This design contrasts legal-corpus adaptation with long-sequence architecture, although the reported experiments do not completely isolate their contributions across all tasks \citep{xiao2021lawformer}. MultiLegalPile, for example, documents its language and jurisdiction coverage and discusses licenses for its subsets \citep{multilegalpilegbmultilingual2023}. Corpus documentation should separately report jurisdiction coverage, collection time, license assumptions, and contamination controls; those are evaluation requirements proposed here rather than properties established for every legal pretraining resource.

\subsection{Instruction Tuning, Synthetic Data, and Specialized Models}

Instruction tuning adapts a model to produce desired legal outputs such as answers, summaries, structured extractions, or draft documents. TransformLLM explicitly uses LLM transformation for domain adaptation, and LawGPT studies knowledge-guided legal-data generation \citep{arbel2024transformllmadaptinglargelanguage,zhou2025lawgptknowledgeguideddatageneration}. These approaches illustrate synthetic or model-assisted supervision but do not establish that all legal instruction-tuning pipelines share the same data lineage. Each system should therefore report how examples were generated, filtered, and validated, and should test whether teacher errors or unrealistic task distributions propagate into the adapted model.

Law-specific generative LLMs include models trained or tuned on legal corpora, regional legal materials, or legal task instructions. Examples include LexGPT, ChatLaw, InternLM-Law, SaulLM, LawGPT-style systems, regional sovereign models, and other legal generative models \citep{lexgpt2023,chatlaw2023,internlmlaw2024,saullm2024,shu2024lawllmlawlargelanguage,zhou2025lawgptknowledgeguideddatageneration,han2025hkgaiv1regionalsovereignlarge,huang2026polilegallmtechnicalreportlarge}. HUKUKBERT is instead a Turkish legal BERT encoder and should be compared with encoder-based legal language models rather than generative legal LLMs \citep{ztrk2026hukukbertdomainspecificlanguagemodel}. These systems reflect different design choices, including continued pretraining, knowledge-base integration, instruction tuning, and model scaling. LawBench and LexEval provide shared task suites for evaluating subsets of this landscape \citep{lawbenchbenchmarkingknowledge2023,lexevalcomprehensivechinese2024}, but direct model comparisons still depend on data, prompting, and evaluation protocol.

These adaptation paths differ in where they place knowledge and control. Continued pretraining stores statistical regularities in model parameters; instruction tuning shapes task formats and response preferences; synthetic data changes the coverage and noise of supervision; specialized models jointly adjust corpora, capacity, and objectives; and model collaboration distributes knowledge and control among components. Parameterized knowledge simplifies individual inference calls but makes updating and tracing an answer to a specific source difficult. External retrieval collections expose source versions while adding retrieval, reranking, and citation-verification costs. The choice depends on update frequency, privacy constraints, interpretability requirements, and computational budget.

\subsection{Model Collaboration and Knowledge Updating}

Recent work also explores collaboration between black-box LLMs and smaller domain-specific models \citep{bladeenhancingblack2024}. This direction is important because legal deployment may face cost, privacy, latency, and audit constraints. A small model can serve as retriever, verifier, classifier, reranker, or domain critic while a large model handles generation. Such hybrid systems complicate evaluation: errors may arise from the small model, the large model, or the interface between them.

A central distinction is between adapting a model to legal language and evaluating it against current authority. Continued pretraining does not by itself document which version of a statute or precedent supports an answer. Work on legal knowledge updating and time-sensitive legal search begins to expose this problem \citep{fan2026llmstimetravelenhancing,prior2026askingoldfrienddiagnosing}. This survey therefore proposes that systems relying on changing law report source versions, retrieval dates, update procedures, and time-specific tests rather than treating domain adaptation as a substitute for updating.

Reproducing an adaptation study requires the corpus size and time range, base model and parameter count, training or adaptation procedure, evaluation data and legal time point, inference prompts, and external updating mechanism. Without these details, an apparent advantage of a domain model may reflect input length, training data, parameter scale, prompt changes, or data overlap rather than legal knowledge alone. Our evidence comparisons retain the datasets, metrics, and protocols used within each original study.

In this review, domain adaptation is one component of the proposed authority-grounded system stack. It targets legal language and task behavior, but adaptation results do not by themselves establish that an answer cites valid authority, reflects current law, or applies to the correct jurisdiction. Retrieval, structured legal metadata, temporal tests, and assurance reporting are therefore separate design and evaluation directions.

\section{Retrieval, Grounding, and Legal Agents}

Domain adaptation can make models more familiar with legal language, but it does not establish which external authority supports a particular answer. Retrieval, citation checking, knowledge representations, and agentic tool use are therefore important system mechanisms examined in this section.

\subsection{Retrieval and Reranking}

Legal RAG systems aim to connect model outputs with statutes, cases, regulations, contracts, or policies. LegalBench-RAG evaluates precise retrieval, while other work studies multi-phase retrieval, knowledge-representation-assisted RAG, claim-level evaluation, Canadian case-law RAG, and privacy-law grounding \citep{goldcoingroundingprivacy2024,enhancingdocumentretrievalmulti2024,kragframeworkenhancingdomain2024,legalbenchragbenchmark2024,das2026finegrainedclaimlevelragbenchmark,zhao2026canlegalragbenchevaluatingretrievalaugmentedgeneration}. These studies motivate, but do not all implement, the broader authority checks proposed here: legal relevance, source status, jurisdiction, valid time, and procedural context.

A legal retrieval pipeline can contain document segmentation, sparse or dense retrieval, reranking, passage selection, prompt construction, answer generation, and citation post-processing. LegalBench-RAG isolates retrieval because end-to-end generation scores cannot diagnose retrieval quality, while LegalAgentBench separately records intermediate progress in tool-using workflows \citep{legalbenchragbenchmark2024,legalagentbenchevaluatingllm2024}. Building on this component separation, we propose reporting stage-specific failures rather than attributing every error to the base model.

We compare RAG, knowledge graphs, and agents through a five-stage workflow: retrieval, reranking, generation, citation verification, and human review. Retrieval determines candidate coverage, reranking prioritizes materials within a limited context, and generation organizes them into an answer. Citation verification checks proposition support and source applicability; human review handles missing facts, conflicting authority, high-risk advice, and escalation. These stages locate the evidence supplied by a study. A study evaluating one stage establishes evidence for that stage, even when its architecture could support a larger workflow.

Errors can propagate across stages. High-recall retrieval may admit low-status or outdated materials, while accurate reranking cannot recover controlling authority absent from the candidate set. A generator can cite a source correctly but misapply the facts, and a citation verifier may establish literal support without determining binding force. Human review also depends on the interface, available time, and information presented. Evaluation should record stage inputs and outputs, failure types, and review cost, using ablations or controlled component replacement to locate the source of performance changes.

\subsection{Generation, Citation Verification, and Knowledge Structures}

A legal answer can fail through a wrong conclusion, a fabricated authority, a misquotation, or a source that does not support the generated proposition. Empirical tool audits, HalluGraph, faithfulness-and-abstention evaluation, and citation-grounding work operationalize different parts of this problem \citep{hallucinationfreeassessingreliability2024,nol2025hallugraphauditablehallucinationdetection,measuringfaithfulnessabstention2025,citationgroundingdetecting2026,whocheckscitations2026}. This survey proposes a stricter reporting requirement: each material legal proposition should be linked to supporting text, while source-status and applicability checks should be reported separately.

Legal texts contain structured relations among citations, holdings, parties, obligations, definitions, procedural events, and time. Recent work explores LLM-driven legal knowledge-graph construction, graph-assisted retrieval, and large-scale citation-graph construction \citep{meher2025corekgllmdrivenknowledgegraph,chen2026legalgraphragmultiagentgraphretrievalaugmented,ovcharov2026automaticconstructionlegalcitation}. Such representations can store relations relevant to authority analysis, but the cited systems do not collectively demonstrate reliable detection of overruling, distinguishing, or binding status. Those relations require dedicated extraction, validation, and update evaluation; graph presence alone should not be treated as symbolic legal enforcement.

RAG and knowledge graphs both use external knowledge but organize and constrain it differently. Flat retrieval can efficiently recall passages from large collections, yet segmentation may disrupt cross-provision references, subsequent case treatment, and conflicting authority. Graphs preserve explicit entity, citation, and temporal relations, while relying on extraction accuracy, relation coverage, and maintenance. A neuro-symbolic case-retrieval study demonstrates the use of case-level and statute-level logical rules for ranking and explanation within its own setting \citep{logicrulesexplanations2024}. This does not establish complete legal reasoning or authority-validity assessment. Both approaches need to report structural coverage, updating failures, and legal relations they do not represent.

\subsection{Agents and Human Review}

Recent work explores multi-agent legal reasoning, court simulation, contract analysis, long-document QA, legal query understanding, time-aware legal agents, and interactive consultation \citep{cangrasptheoriesenhance2024,agentscourtbuildingjudicial2024,multiagentsimulator2025,legalagentbenchevaluatingllm2024,agentslawtaxonomyapplications2026,chen2026legalgraphragmultiagentgraphretrievalaugmented,raptopoulos2025paktonmultiagentframeworkquestion,li2026legalmalrmultiagentqueryunderstandingllmbased,fan2026llmstimetravelenhancing}. The implemented tool calls and decompositions differ by study. This motivates direct tests for intermediate error propagation, hidden assumptions, structured-output failure, tool misuse, and responsibility rather than assuming either benefit or reliability from the agent architecture alone.

LegalAgentBench demonstrates the value of measuring intermediate progress rather than only final success, and the legal-agent survey catalogs architectures and application settings \citep{legalagentbenchevaluatingllm2024,agentslawtaxonomyapplications2026}. Building on this evidence, we propose that deployment-oriented evaluations record tool calls, retrieved sources, assumptions, uncertainty, and human interventions. These are prospective design requirements; the cited agent studies do not establish that logging or checkpoints alone improve professional outcomes.

Agents turn this workflow into a control process that can iterate and invoke tools. They enable re-retrieval, decomposition, and verification triggers while making system state and responsibility boundaries harder to specify. LegalAgentBench records final success and intermediate progress across 17 corpora, 37 tools, and 300 tasks \citep{legalagentbenchevaluatingllm2024}. Tool success in this environment does not establish reduced risk in lawyers', judges', or institutional workflows. Human review in actual use must specify who can override the model, which failures trigger escalation, which sources and intermediate states reviewers can inspect, and how their decisions are recorded.

In the proposed taxonomy, retrieval, grounding, and agents form a mechanism layer: retrieval supplies candidate sources, grounding tests generated claims against evidence, and agents organize multi-step workflows. Whether the integrated system preserves component-level performance is an empirical question; evaluations should therefore report source mismatch, intermediate assumptions, and unverified steps where relevant.

\section{Benchmarks and Evaluation}

Legal LLM evaluation has expanded from task accuracy to multi-dimensional assessment. Benchmarks and evaluation studies now examine legal knowledge, reasoning, retrieval, RAG, agent behavior, multilingual QA, temporal event reasoning, embeddings, compliance, safety, fairness, and open-ended text quality \citep{lawbenchbenchmarkingknowledge2023,lexevalcomprehensivechinese2024,legalagentbenchevaluatingllm2024,lextimebenchmarktemporalordering2025,legalevalqnew2025,butler2025massivelegalembeddingbenchmark,wang2025llmbasedhsecomplianceassessment,hui2025tridentbenchmarkingllmsafety,barale2025fairnessisntstatisticallimits,multibenchevaluating2026}. Their units of analysis and evidential claims differ, so the label ``benchmark'' should not be used interchangeably with a fairness audit, system study, or deployment evaluation.

\subsection{Evaluation Dimensions and Benchmark Families}

\begin{table}[!htbp]
\centering
\footnotesize
\caption{Assurance dimensions, evaluation targets, and typical failure modes for legal AI systems.}
\label{tab:evaluation-dimensions}
\begin{adjustbox}{max width=\linewidth}
\begin{tabular}{L{0.22\linewidth} L{0.34\linewidth} L{0.34\linewidth}}
\toprule
Dimension & What it tests & Typical failure mode \\
\midrule
Task accuracy & Correctness on classification, retrieval, QA, or generation tasks & Benchmark artifacts and shallow pattern matching \\
Grounding & Whether outputs are supported by cited legal sources & Fabricated citations or misquoted authority \\
Jurisdictional robustness & Whether answers remain valid under different legal systems & Transferring rules across jurisdictions \\
Temporal validity & Whether current law and precedent are used & Reliance on outdated authority \\
Fairness and safety & Whether outputs create disparate or unsafe effects & Biased reasoning or overconfident advice \\
Professional usability & Whether legal users can inspect and rely on the system & Plausible but unauditable reasoning \\
\bottomrule
\end{tabular}
\end{adjustbox}
\end{table}

Task accuracy is useful when the output space is well defined, such as charge prediction, article prediction, or multiple-choice legal knowledge. For open-ended answers, memos, summaries, or multi-step workflows, this review proposes additional checks for source support, jurisdictional and temporal scope, and usefulness to the intended user. The appropriate combination of automatic metrics, source verification, expert review, and task-specific rubrics remains an evaluation question rather than a settled standard.

Existing benchmarks differ most importantly in their unit of evaluation. Capability-oriented suites measure legal knowledge and the application of rules to facts, including exceptions and multi-step reasoning \citep{lawbenchbenchmarkingknowledge2023,lexevalcomprehensivechinese2024,dai2025laiw,liu2025lexgeniusexpertlevelbenchmarklarge,syllogismpromptingteachingjudgment2023,le2026benchmarkingreasoningdualaspectlargescale}. Component-oriented benchmarks isolate retrieval, embeddings, claim-level generation, or intermediate agent behavior \citep{legalbenchragbenchmark2024,butler2025massivelegalembeddingbenchmark,das2026finegrainedclaimlevelragbenchmark,zhao2026canlegalragbenchevaluatingretrievalaugmentedgeneration,legalagentbenchevaluatingllm2024}. Output-oriented evaluations cover judgment or document generation, open-ended quality, and case-based faithfulness or abstention \citep{judgebenchmarkingjudgmentdocument2025,legalevalqnew2025,measuringfaithfulnessabstention2025}. Other studies vary language, jurisdiction, legal tradition, or event temporality \citep{whichinstitutionalframeworksdo2026,lextimebenchmarktemporalordering2025,multibenchevaluating2026}. LexTime specifically evaluates ordering of legal events; it does not test whether a system retrieves the legally valid version of a rule.

Table~\ref{tab:benchmark-landscape} compares representative datasets and benchmarks for Chinese legal AI. CAIL2018 and the LeCaRD series define tasks and supervision signals. LawBench, LexEval, and LAiW organize multiple tasks around different cognitive or legal-reasoning structures. LegalAgentBench adds tools and intermediate progress to the unit of evaluation. Because the tasks, candidate spaces, output formats, and metrics differ, the table records scale, protocol, availability, and limitations rather than comparing model scores across benchmarks.

\begin{table}[!htbp]
\centering
\footnotesize
\setlength{\tabcolsep}{3pt}
\renewcommand{\arraystretch}{1.12}
\caption{Representative Chinese legal AI datasets and benchmarks. All use Chinese legal materials or scenarios. The source/data column identifies the reported release or collection; it does not imply that each benchmark tests legal validity at a specified date.}
\label{tab:benchmark-landscape}
\begin{tabularx}{\linewidth}{@{}L{0.12\linewidth} L{0.15\linewidth} L{0.16\linewidth} L{0.13\linewidth} Y Y@{}}
\toprule
Dataset & Main task & Scale and scope & Source/data & Metrics and availability & Main limitations \\
\midrule
CAIL2018 \citep{xiao2018cail2018} & Article, charge, and sentence prediction & Chinese criminal law; over 2.6 million cases & Original dataset release & Classification and sentencing metrics; public data and baselines & Criminal prediction; labels extracted from document structure; not open-ended reasoning \\
LeCaRD \citep{ma2021lecard} & Similar-case retrieval & Chinese criminal law; 107 queries, 10\,716 candidates & Original dataset release & Ranking metrics; public data & Small query set and candidate pool; limited relevance and collection coverage \\
LeCaRDv2 \citep{li2024lecardv2} & Large-scale case retrieval & Chinese criminal law; 800 queries, 55\,192 candidates & Built from 4.3 million judgments & Ranking metrics; multiple legal annotators & Fixed candidate pool; offense, sentencing, and procedural relevance do not establish authority status \\
LawBench \citep{lawbenchbenchmarkingknowledge2023} & Legal knowledge, understanding, and application & Chinese law; 20 tasks, approximately 10\,000 questions & Original benchmark release & Task-specific metrics; public code, data, and predictions & Heterogeneous sources and formats; aggregate score obscures specific failures \\
LexEval \citep{lexevalcomprehensivechinese2024} & Legal knowledge, reasoning, application, and ethics & Chinese law; 23 tasks, 14\,150 questions & Original benchmark release & Task-specific metrics; public data and leaderboard & Potential contamination; limited regional, document, and legal-update coverage \\
LAiW \citep{dai2025laiw} & Syllogism-oriented legal evaluation & Chinese law; three capability levels and multiple task types & Original benchmark release & Automatic evaluation with legal-expert review & Levels use different scales; human-review samples and roles limit external validity \\
LegalAgent\newline Bench \citep{legalagentbenchevaluatingllm2024} & Tool use and multi-step tasks & Chinese legal scenarios; 17 corpora, 37 tools, 300 tasks & Original benchmark release & Success and intermediate progress; public code and data & Simulated tools do not reproduce institutional permissions, systems, or professional responsibility \\
\bottomrule
\end{tabularx}
\end{table}

\subsection{Validity, Failure Modes, and Reporting}

High benchmark scores do not by themselves establish professional usefulness because most benchmarks hold users, interfaces, operating policies, and responsibility structures fixed or outside scope. Separate studies show citation hallucination in legal research tools and failures to abstain in case-based argument generation \citep{hallucinationfreeassessingreliability2024,measuringfaithfulnessabstention2025}. We therefore recommend that deployment-oriented evaluations add source applicability, abstention, time-specific tests, and human-review burden when those properties are material to the intended use.

Benchmark validity has at least four dimensions. Construct validity concerns whether the test measures legal reasoning or vocabulary and templates. Internal validity concerns consistent control of prompts, randomness, and scoring. External validity concerns transfer to other courts, regions, document types, or users. Consequential validity concerns whether optimizing the benchmark encourages unsafe behavior. Improved accuracy on a fixed question set does not directly demonstrate reliability when facts are incomplete, rules conflict, or the governing law changes.

LLM-based evaluators can scale assessment of open-ended outputs, but agreement with a model-based rubric does not independently establish legal correctness. For high-stakes settings, we recommend using such evaluators as screening or aggregation tools alongside deterministic source checks and expert review, with evaluator prompts, models, repeat runs, and human-agreement evidence reported explicitly.

Public legal texts and exam questions may overlap with model training data, making contamination controls important for capability claims. Legal applicability can also change with amendments and later cases. We therefore propose that benchmark documentation specify the source corpus, jurisdiction, legal time point, contamination checks, and update policy. Time-specific legal validity should be evaluated directly rather than inferred from event-ordering or general temporal-reasoning scores.

\subsection{System Evidence Packages}

For legal LLM systems, we propose accompanying aggregate scores with a report that identifies the evaluated system boundary and records task accuracy, retrieval quality, citation support, jurisdictional and temporal scope, abstention policy, fairness or privacy tests where relevant, and human-review cost. This is a reporting agenda derived from the synthesis, not an existing consensus standard.

Evaluation is where the survey's proposed system perspective becomes testable. A model-centric leaderboard provides evidence about the measured tasks, but not automatically about an unmeasured workflow or operating condition. We therefore propose evaluation packages that connect task performance with source faithfulness, temporal and jurisdictional tests, abstention behavior, and human-review burden where those properties are material to the intended use.

\section{Risks, Governance, and Responsible Deployment}

Legal LLM governance cannot be reduced to a generic list of model risks. The relevant control depends on the system boundary, the person or institution using the output, the governing legal regime, and the consequence of error. Empirical legal studies document citation hallucination, failures of faithfulness or abstention, bias in case summaries, and limitations of statistical fairness criteria \citep{hallucinationfreeassessingreliability2024,measuringfaithfulnessabstention2025,questioningbiasescase2023,barale2025fairnessisntstatisticallimits}. These studies support particular risk claims; they do not prove that every legal LLM exhibits every failure.

\subsection{Models, Corpora, and Retrieval Sources}

\begin{table}[!htbp]
\centering
\footnotesize
\caption{Failure modes, proposed controls, and observable assurance signals. The controls are research and governance directions; each requires direct evidence that it reduces the stated failure in the intended setting.}
\label{tab:risk-mitigation}
\begin{adjustbox}{max width=\linewidth}
\begin{tabular}{L{0.18\linewidth} L{0.28\linewidth} L{0.28\linewidth} L{0.16\linewidth}}
\toprule
Risk & Legal consequence & Mitigation direction & Evaluation signal \\
\midrule
Hallucinated authority & cites non-existent or unsupported law & citation-existence and proposition-support checks & citation precision and support rate \\
Outdated law & applies superseded statutes or cases & source versions and time-specific retrieval tests & validity at the stated legal time point \\
Jurisdiction mismatch & transfers rules across legal systems & jurisdiction metadata and retrieval filters & scope-error rate across jurisdictions \\
Bias and unfairness & biased summaries, classifications, or advice & task-specific scenario and counterfactual audits & subgroup and scenario error analysis \\
Confidentiality failure & discloses client or restricted information & input policy, access control, and vendor review & leakage tests and incident records \\
Over-reliance & users accept plausible but wrong outputs & role-specific review and escalation policy & override, correction, and review burden \\
Opaque workflow & errors cannot be traced or corrected & versioned source and action records & trace coverage and incident resolution time \\
\bottomrule
\end{tabular}
\end{adjustbox}
\end{table}

Governance begins before inference. A system report should identify the base model and provider, model version, training or adaptation data that can be disclosed, jurisdiction and language coverage, collection dates, license assumptions, and known access restrictions. MultiLegalPile illustrates how a legal corpus can report languages, jurisdictions, sources, and subset licenses \citep{multilegalpilegbmultilingual2023}; it does not resolve the licensing or privacy status of every downstream model. The NIST Generative AI Profile provides a cross-sector risk-management resource for incorporating trustworthiness considerations across design, development, use, and evaluation \citep{autio2024nistgenaiprofile}. We use it as general lifecycle guidance, not as evidence that a legal system is compliant with a particular jurisdiction.

Retrieval governance concerns which collections may be searched, how source versions are tracked, how conflicts are surfaced, and whether sensitive or restricted materials can enter prompts. Legal research-tool audits show that proprietary RAG-based systems can still produce hallucinated or inaccurate answers \citep{hallucinationfreeassessingreliability2024}. A retrieval log therefore does not establish correctness by itself. For the proposed framework, a deployment record should distinguish document provenance, retrieval relevance, proposition support, legal status, jurisdiction, and valid time, and should preserve failed retrievals or conflicting authority for review.

\subsection{Professional Responsibility and Institutional Compliance}

Professional obligations cannot be delegated to a model. ABA Formal Opinion 512 identifies duties relevant to lawyers using generative AI, including competence, confidentiality, communication, supervision, candor, and reasonable fees \citep{aba2024formalopinion512}. The opinion is U.S.-oriented professional guidance and should not be generalized to all legal systems. Its practical implication for system design is narrower: the responsible professional needs enough information about model limitations, data handling, and output verification to discharge the duties that apply in the relevant jurisdiction. Human review should therefore be specified by role, trigger, and decision authority rather than mentioned as an unspecified safeguard.

Applicable obligations depend on the actor and intended use. The EU AI Act classifies certain systems used by or on behalf of judicial authorities to research and interpret facts and law, or apply law to facts, as high-risk while excluding some purely ancillary administrative uses \citep{eu2024aiact}. This is an intended-use distinction, not a rule that every legal chatbot is high-risk. Institutional deployment should therefore document the legal basis for use, affected users, decision authority, procurement and vendor responsibilities, required impact or conformity assessments, and routes for contesting or correcting outputs.

\subsubsection{The Chinese Legal and Regulatory Context}
In China, the actor, intended users, and purpose of a system also determine which obligations apply. The \emph{Interim Measures for the Management of Generative Artificial Intelligence Services} apply to services that use generative AI to provide generated content to the public within the People's Republic of China. They expressly exclude research, development, and application activities that do not provide such services to the domestic public. The measures address lawful sourcing and quality of training data, personal information, complaints, and, where applicable, security assessments and algorithm filing.\footnote{Cyberspace Administration of China and other authorities, \emph{Interim Measures for the Management of Generative Artificial Intelligence Services}, 13 July 2023: \url{https://www.cac.gov.cn/2023-07/13/c_1690898327029107.htm}.} Research prototypes, internal institutional assistants, and public-facing legal QA services therefore cannot be assigned the same regulatory category by default. Sector-specific rules and professional obligations in judicial, lawyer, and public legal services must be assessed separately.

Data and security governance also involve the \emph{Personal Information Protection Law of the People's Republic of China}\footnote{\emph{Personal Information Protection Law of the People's Republic of China}, 20 August 2021: \url{https://www.cac.gov.cn/2021-08/20/c_1631050028355286.htm}.} and GB/T 45654--2025, \emph{Cybersecurity Technology---Basic Security Requirements for Generative Artificial Intelligence Services}.\footnote{National Technical Committee 260 on Cybersecurity, GB/T 45654--2025, 30 June 2025: \url{https://www.tc260.org.cn/portal/article/1/20250630122232}.} These instruments inform checks on data sources, personal information, model and corpus versions, content security, and incident handling. Citing an instrument does not establish compliance: a system report must specify how its requirements map to components, who implements them, and what tests supply evidence that the controls work.

\subsection{Bias, Privacy, and Post-Deployment Monitoring}

Bias claims should be tied to the evaluated legal task. Questioning Biases studies case-judgment summaries and their source datasets, while Barale's work analyzes limits of statistical fairness in legal LLM evaluation \citep{questioningbiasescase2023,barale2025fairnessisntstatisticallimits}. General-domain bias benchmarks can motivate hypotheses but do not directly establish biased legal summaries or judgments. Privacy and confidentiality likewise require legal-context evidence. Legal NLP work discusses data-protection questions around model training, and ABA Formal Opinion 512 requires lawyers to evaluate disclosure risks before entering client information into generative AI tools \citep{almeida2024legalframeworknaturallanguage,aba2024formalopinion512}. General multimodal privacy benchmarks are best treated as cross-domain technical evidence rather than proof of legal confidentiality compliance.

Model, corpus, law, and user behavior can change after an initial evaluation. We therefore propose a post-deployment record containing model and retrieval-corpus versions, material prompt or policy changes, observed citation and scope failures, overrides and escalations, user complaints, correction time, and criteria for suspension or rollback. NIST's lifecycle framing supports ongoing risk management in general \citep{autio2024nistgenaiprofile}; the specific incident fields above are the survey's proposed legal-system reporting schema and require prospective validation.

Each governance layer carries a different evidence responsibility. The model and corpus layer records data documentation, licenses, and versions. The retrieval-source layer records source inventories, updates, and conflict handling. The professional layer needs review protocols, competence boundaries, and communication records; the institutional layer specifies intended use, permissions, and impact assessments; and the post-deployment layer records monitoring measures, incidents, and suspension criteria. These forms of evidence are not interchangeable. Passing a safety benchmark does not establish that a retrieval collection was lawfully sourced. Requiring human review does not establish that reviewers have adequate information, and user warnings cannot replace technical source verification.

No single control establishes legal reliability. HalluGraph, for example, supports auditable hallucination detection, but it does not by itself establish jurisdictional applicability, confidentiality, or professional fitness \citep{nol2025hallugraphauditablehallucinationdetection}. Source grounding, abstention, structured representations, human review, privacy controls, and monitoring should therefore be treated as hypotheses about particular failure modes. Each control needs an evaluation showing which failure it reduces, under which operating conditions, and what new failure or review burden it introduces.

The five governance layers assign different responsibilities and require different evidence: model and corpus governance, retrieval sources, professional workflows, institutional compliance, and post-deployment monitoring. This layered view avoids treating a benchmark score, a RAG architecture, a user warning, or a policy document as a complete governance solution. It also makes the proposed authority-grounded framework auditable: a system report can reveal which layer was not evaluated or which responsibility remains unassigned.

\section{Open Challenges and Research Agenda}

The literature reviewed above motivates a research agenda that treats legal authority, uncertainty, and human review as explicit system requirements. The directions below are proposals synthesized by this review; they are not presented as empirically validated design standards or as deficiencies measured across the field. We organize them by the assurance obligation that future work could make testable. Table~\ref{tab:research-agenda} summarizes the proposed directions and their evaluation requirements.

\subsection{Traceability and Temporal and Jurisdictional Robustness}

We propose that future systems expose which legal sources, factual assumptions, and interpretive steps support an answer. A trace should be checkable by legal professionals rather than merely plausible to a language model. Existing work provides evidence for evaluating retrieval stages, citation support, and abstention \citep{legalbenchragbenchmark2024,measuringfaithfulnessabstention2025,citationgroundingdetecting2026}; linking every material proposition to applicable authority and separating retrieved text from model interpretation remain broader design and evaluation goals.

Legal authority changes over time and differs across jurisdictions. LexTime evaluates temporal ordering of events rather than legal-source version control \citep{lextimebenchmarktemporalordering2025}, while time-aware legal search and institutional analyses illustrate narrower aspects of temporal and jurisdictional scope \citep{whichinstitutionalframeworksdo2026,fan2026llmstimetravelenhancing}. We therefore propose evaluations that hold the legal question constant while varying the relevant date and jurisdiction, alongside explicit temporal metadata, retrieval filters, and source-version records. These mechanisms and tests should be evaluated directly rather than assumed to prevent cross-jurisdiction or outdated-law errors.

\begin{table}[!htbp]
\centering
\footnotesize
\caption{Proposed research agenda synthesized by this review. The directions and evaluation requirements are design proposals to be tested, not empirically validated standards.}
\label{tab:research-agenda}
\begin{adjustbox}{max width=\linewidth}
\begin{tabular}{L{0.20\linewidth} L{0.24\linewidth} L{0.24\linewidth} L{0.22\linewidth}}
\toprule
Challenge & System direction & Evaluation requirement & Deployment risk addressed \\
\midrule
Traceable reasoning & source-linked reasoning traces and citation graphs & proposition-level source support & hallucinated or misquoted authority \\
Temporal robustness & time-aware retrieval and source versioning & time-sliced legal tests & outdated statutes or precedents \\
Jurisdictional scope & jurisdiction-aware prompts, retrieval filters, and metadata & cross-jurisdiction transfer tests & applying the wrong legal system \\
Workflow reliability & agent checkpoints, audit logs, and human review & step-level success and intervention cost & compounding multi-step errors \\
Data governance & provenance, licensing, and privacy controls & corpus documentation and leakage tests & privacy and licensing violations \\
Open-ended generation & task-specific rubrics and expert review & faithfulness, completeness, and abstention & fluent but unusable legal drafts \\
Evidence-package reporting & reports tied to system operating conditions & multi-metric system reports & leaderboard scores hiding operating-condition risks \\
\bottomrule
\end{tabular}
\end{adjustbox}
\end{table}

\subsection{Dynamic Evaluation and Human-AI Workflows}

Current benchmarks provide components for temporal ordering, agent-step evaluation, and claim-level retrieval analysis \citep{lextimebenchmarktemporalordering2025,legalagentbenchevaluatingllm2024,das2026finegrainedclaimlevelragbenchmark}. Building on them, we propose controlled evaluations that vary law, facts, and user goals, and that report abstention, evidence support, retrieval failure, and expert-review burden in addition to task accuracy. The utility and reliability of these proposed measurements require validation with legal professionals.

Legal LLMs should be studied as components in human workflows. The key question is not whether a model can replace a lawyer, but which parts of legal work can be assisted safely, transparently, and accountably. Agent benchmarks and taxonomies already expose the need to evaluate intermediate actions and tool use, but user-specific intervention cost remains underdeveloped \citep{legalagentbenchevaluatingllm2024,agentslawtaxonomyapplications2026}. Different users need different safeguards: a professional may need source and decision records, while a lay user may need warnings, clarification questions, and referrals.

\subsection{Constraints, Data Governance, and Reporting}

Legal reasoning often has explicit rule structure, but full formalization of law is difficult. Existing work explores logic-based explanations and knowledge-representation-assisted retrieval \citep{logicrulesexplanations2024,kragframeworkenhancingdomain2024}; the latter does not by itself establish symbolic enforcement of legal rules. A research direction is to test hybrid components for rule checking, obligation tracking, temporal constraints, or citation validation, while measuring formal coverage and documenting where interpretive judgment remains necessary.

Legal corpora raise licensing, privacy, representativeness, and update issues \citep{multilegalpilegbmultilingual2023,almeida2024legalframeworknaturallanguage,java2024operationalizingrightdataprotection}. Future work should report corpus provenance, jurisdiction coverage, time range, licensing assumptions, preprocessing decisions, and whether sensitive information is included. Reproducibility is especially important because legal datasets may be removed, modified, or restricted over time.

For deployment-oriented research, we propose supplementing model leaderboards with evidence packages that report data provenance, retrieval coverage, citation faithfulness, robustness tests, privacy analysis, interface safeguards, human-review protocols, and post-deployment monitoring. This is a research direction rather than a claim that a legal certification regime already exists or that this package has been validated: the immediate goal is to make system claims comparable, inspectable, and tied to operating conditions.

To make these directions falsifiable, we describe them through an evidence gap, a research question, a candidate mechanism, and a validation requirement. The gap must follow from reviewed evidence or an explicit methodological limitation. The question identifies the task, jurisdiction, time, and users; the mechanism remains a candidate explanation. Validation requires a comparison condition, failure measures, a human-review protocol, and a stated boundary of external validity. For temporal failures, a concrete study could compare version filters, temporal indexes, and re-retrieval around a specified legislative amendment, measuring outdated-source use alongside missed evidence, incorrect abstention, and review cost.

\section{Limitations}

The reviewed studies do not support a single cross-paper ranking of legal LLM systems. Tasks, jurisdictions, languages, corpora, model access conditions, metrics, legal time points, and system boundaries differ substantially. The comparisons in this survey therefore identify recurring mechanisms and evidence gaps; they do not treat results obtained under incompatible settings as entries on one leaderboard.

Most reported evidence comes from benchmarks, prototypes, and offline experiments. These studies can establish task behavior or component performance under stated conditions, but they rarely measure longitudinal use in professional workflows, institutional consequences, recovery from error, expert-review burden, or post-deployment monitoring. The literature therefore does not yet support general claims about deployment effectiveness across legal roles and organizations.

Reporting of legal authority is also inconsistent across research threads. Studies differ in whether they record source hierarchy, binding force, jurisdiction, valid time, procedural posture, and proposition-level support. This limits synthesis across retrieval, citation, generation, and agent studies even when each component is evaluated carefully within its own task.

Finally, the authority-grounded system remains a proposed organizing and evaluation framework rather than a validated deployment standard. The structured evidence map was screened and coded by one reviewer at title-and-abstract level, so its counts support process reproduction and literature navigation rather than field prevalence estimates. The reviewed component evidence does not collectively establish that the full framework improves legal outcomes; prospective system evaluations and human-centered studies remain necessary.

\section{Conclusion}
Legal LLM research concerns both whether a model can answer an individual task and whether the surrounding system supplies a basis for reliance. The system needs to identify potentially applicable authority, preserve jurisdictional and temporal scope, connect material claims to evidence, expose uncertainty and intermediate decisions, and locate responsibility within a human workflow. We organize this problem along the linked axes of task capability, system mechanisms, and assurance evidence, and define \emph{authority grounding} as a system integration contract with requirements beyond RAG, citations, or agents alone.

The reviewed literature evaluates capability, retrieval quality, citation support, agent behavior, and governance at different system boundaries. Research reports should therefore state where authority enters the pipeline, how source applicability and claim support are tested, when systems abstain or escalate, and who remains responsible for use. Certification-style reports are a research direction, not an existing legal certification regime. Progress requires direct evidence that specific controls reduce specific failures under stated jurisdictions, legal time points, users, and operating conditions.

\section*{Data and Materials Availability}
The supplementary materials distributed with this manuscript include the search manifest, screening protocol, all 381 screening decisions, the 327-record evidence map, assurance-review decisions, exploratory figure values, representative-study evidence, and publication-status records. The accompanying literature index is available at \url{https://github.com/Jeryi-Sun/LLM-and-Law}. The supplementary records document the frozen review sample, while the literature index may continue to change.

\section*{Acknowledgments}
This work was supported by the National Key Research and Development Program of China (2023YFA1008704) and the National Natural Science Foundation of China (62472426 and 62502091).

\bibliography{main}
\end{document}